\documentclass{sig-alternate}
\usepackage{url}

\newcommand{\abs}[1]{\left|#1 \right|}
\newcommand{\paren}[1]{\left( #1 \right)}

\newcommand{\comment}[1]{}

\begin{document}
%
\conferenceinfo{WOODSTOCK}{'97 El Paso, Texas USA}



\title{OFF-Set: One-pass Factorization of Feature Sets for Online Recommendation in Persistent Cold Start Settings}

%
%
%
%
%

\numberofauthors{2} 
%
\author{
%
%
\alignauthor
Michal Aharon, Natalie Aizenberg, Edward Bortnikov, Ronny Lempel \\
\affaddr{Yahoo! Labs, Haifa, Israel}\\
\affaddr{\{michala,anatalia,ebortnic,rlempel\}@yahoo-inc.com}
\and
\alignauthor
Roi Adadi, Tomer Benyamini, Liron Levin, Ran Roth, Ohad Serfaty\\
\affaddr{Yahoo! Smart Ad, Tel Aviv, Israel}\\
\affaddr{\{roiaddi,tomerb,liron,ranr,ohad\}@yahoo-inc.com}\\
}
\date{30 July 1999}

\maketitle
\begin{abstract}
One of the most challenging recommendation tasks is recommending to a new,
previously unseen user. This is known as the {\em user cold start} problem.
Assuming certain features or attributes of users are known, one approach for
handling new users is to initially model them based on their features.

Motivated by an ad targeting application, this paper describes an extreme online
recommendation setting where the cold start problem is perpetual.
Every user is encountered by the system just once, receives a recommendation,
and either consumes or ignores it, registering a binary reward.

We introduce One-pass Factorization of Feature Sets, 
{\em OFF-Set}, a novel recommendation algorithm based on Latent
Factor analysis, which models users by mapping their features to a latent space.
Furthermore, OFF-Set is able to model non-linear interactions between pairs of
features. OFF-Set is designed for purely online recommendation, performing lightweight
updates of its model per each recommendation-reward observation.
We evaluate OFF-Set against several state of the art baselines, 
and demonstrate its superiority on real ad-targeting data.
\end{abstract}



\pagebreak

\section{Introduction}

In recent years, Collaborative Filtering (CF) based recommender systems have
gained both commercial success and increased focus from the research
community~\cite{Koren091the}. Generally speaking, CF discovers and exploits
recurring consumption patterns of items by users, at scale, in order to recommend
items to users. Studied consumption types include ratings of items by users,
which explicitly articulate users' likings and dislikings of items; binary
indication of users' consumption of items, without explicit evidence regarding
the users' opinions on their consumed items; richer implicit interaction
settings where users, while still not explicitly providing ratings on items, can
perform multiple operations on items (e.g. click or comment on news stories), etc.

In order to provide high-quality recommendations of items to a user, recommender
systems must observe some past activity of the user. When new users first
arrive to a system, no previous activity is available and they are said to be
``cold''; recommending to such users is challenging, and has been coined the
``user cold-start problem''~\cite{interviewNewUsers,coldStartWithFeatures,representFeatures}.

Motivated by an online ad targeting application (Section~\ref{sec:dynacamp}), 
this paper addresses an extreme online
recommendation setting where the cold start problem is perpetual - practically
every user encountered by the system is seen just once, i.e. every recommendation
request is for a previously unseen user. Nevertheless, certain attributes of the
users are known, and those are leveraged for recommendation. For every
recommendation made, a binary indication (i.e. reward) is observed. The goal is
to maximize the reward.

We introduce One-pass Factorization of Feature Sets - {\em OFF-Set} -- a novel
recommendation algorithm based on {\em latent factor analysis}~\cite{korenMF}, 
which models both users and items by mapping their features to a latent space. 
With OFF-Set we offer a non-linear solution where a latent space is designed in a manner that can allow modeling features both separably and combined.
OFF-Set is formulated for purely online recommendation, performing
lightweight updates of its model per each recommendation-reward observation.
Nevertheless, OFF-Set can leverage training data from any source as its convergence does 
not rely on any specific exploration scheme -- it can learn from user
interactions either with its own recommendations, or with those made by any 
other offline or online scheme. 

We evaluate OFF-Set against several state of the art baselines, on both
synthetic and real-life data. OFF-Set is comparable to the strongest baselines
on synthetic data, and outperforms all baselines on real ad-targeting data.

While our exposition describes OFF-Set in an ad targeting context, the
solution we design is fairly general and may be applied to other real life
problems. One such example is an extreme item cold-start recommendation setting
in e-commerce, where the users may be well known to the system but products
tend to be highly dynamic~\cite{Han_abstractfeature_based}.\\

The contributions of this work are as follows:
\begin{enumerate}

\item OFF-Set exploits user features, and hence is able to offer recommendations
to previously unseen users. Furthermore, it is a latent factor approach
that models dependencies between sets (i.e. combinations) of user features on the
one hand and the possible items on the other hand.

\item OFF-Set is a fully online, single pass algorithm. It performs a single
lightweight update of its model upon every observed recommendation-reward pair
(see Section~\ref{subsec:online}).
\item OFF-Set outperforms state-of-the-art baselines on a real-life
ad-targeting recommendation scenario.
\end{enumerate}	

The rest of this paper is organized as follows. Section~\ref{sec:related}
surveys related work. Section~\ref{sec:dynacamp} describes the problem setting
of dynamic ad campaign optimization, and Section~\ref{problemDescription} formalizes
the problem with some notations.
Section~\ref{solution} presents the OFF-Set algorithm's cost function and general solution, while the details about the latent space design and construction are described in Section~\ref{userCombination}.
Section~\ref{results} reports on our experiments. We conclude our work in
Section~\ref{sec:conc}.

\section{Related Work}
\label{sec:related}
The ever-growing demand for automated recommendations at scale in the recent decade 
has been promoting the development of a great variety of techniques to power recommender systems.
Collaborative Filtering (CF) algorithms are often the alternative of choice, among which, the {\em matrix factorization\/} (MF)~\cite{korenMF}
technique is very popular and successful one ~\cite{MFExample1,MFExample2,radioActive,C2C-WWW2012}.
In its most basic form, MF associates each user and each item with a latent factor (vector). 
The match score between a user and an item is represented as an inner product between the 
corresponding vectors. 
In this context, two users are close in the latent space if their choices overlap
over multiple items. Respectively, two items are close if they were enjoyed by 
many common users. Matrix factorization is beneficial if the original (user,
item) relationship can be closely captured by a low-dimension latent representation.

Although collaborative filtering algorithms have been gaining great success in recent years, 
they are often challenged by the well-known {\em user
cold-start\/} problem~\cite{coldStartWithFeatures} -- namely, making recommendations to
previously unobserved users. Some
works \cite{interviewNewUsers,golbandi:bootstrapping} jump-start the modeling
process by interviewing the new users. However, in highly dynamic environments
like online advertising this approach is not-applicable.  
Other approaches initialize new users' factors based on the overlap of 
their features with those of previously observed users or items. Gantner et
al. exploit these features directly~\cite{coldStartWithFeatures}, while Park and
Chu derive the missing parts through regression~\cite{representFeatures}.
OFF-Set represents users by mapping their features -- along with dependencies
between sets of features -- into a latent space (see
Section~\ref{userCombination}).

Another approach that can capture features well is 
\textit{factorization machines\/} (FM), first presented by Rendle~\cite{FM,FM2}. FM is known for its capability to model sparse feature spaces. It is a combination of {\em support vector machines\/} 
(SVM)~\cite{SVMBook} and matrix factorization, and it closely fits our problem specification. 
FM was shown to be superior to other approaches, including ``regular'' matrix
factorization~\cite{FMLibFM,FMGerman}.

Outside of collaborative filtering there may be found other solutions for feature-based prediction modeling to offer a different approach to the cold-start and sparsity issues.
The \textit{gradient-boosted decision trees\/}(GBDT) algorithm~\cite{GBDT2,GBDT1} 
is one such approach. In contrast with 
linear methods, GBDT models the scoring process as an aggregation over an
ensemble of decision trees. GBDT has been shown to be successful in non-latent feature spaces,
including those with interdepdendent features~\cite{GBDT1}.

Both FM and GBDT are not designed to work in an online setting, as they require
several passes over the data in order to converge. In this sense they differ from 
our approach, as the suggested OFF-Set, is completely online 
-- i.e., every input element is processed exactly once and the ensuing
update of the model is lightweight.

The setting of our problem is similar to that of the contextual multi-armed
bandits framework \cite{langford:contextualbandits}. The context corresponds to
the users' features, and the possible actions are the ads to show the user.
Contextual bandit schemes were used in the past in ad serving contexts (e.g.
\cite{pal:contextualMAB}).
However, while OFF-Set and the baselines we test can be readily evaluated on
offline data, offline evaluation of bandit based algorithms requires rejection
sampling of large fractions of the data and will result in incomparable results
\cite{lihongli:contextualbandits}.

\section{Dynamic Advertising Campaigns}
\label{sec:dynacamp}
Dynamic advertising campaigns are a relatively new form of digital display
advertising, where the advertiser offers multiple products or multiple ad
creative alternatives (referred to as `ad variants') to be presented to
users. 
Unlike traditional display advertising settings, where the ad exchange assigns
a specific ad to every user impression, in dynamic campaigns the exchange
designates a campaign to the impression. It is up to a follow-up process - a
selection algorithm - to determine which variant, i.e. which specific ad
instance, to actually serve to the user. That decision is based upon data
(context) provided to the selection algorithm, e.g. properties about the user,
the page content surrounding the ad slot, the time of day, etc.
In this work we assume only basic demographic information about the users (age,
gender, geographic location) is known.

\begin{figure}
\centering
\epsfig{file=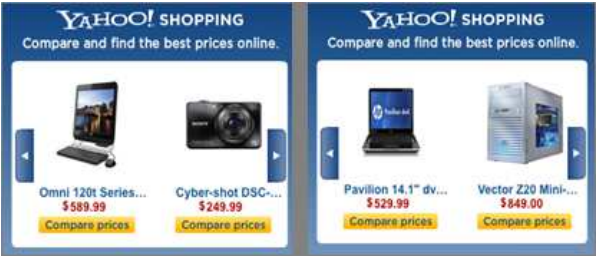, width=0.4\columnwidth,angle=270}
\caption{Example of Dynamic Ad Campaign (for Yahoo! Shopping)\label{exampleYS}}
\end{figure}

A typical optimization task in dynamic campaigns
is to maximize the campaign's click-through rate (CTR), i.e. the CTR over all
user impressions which were served by some ad variant of the campaign.
We formulate dynamic ad campaign optimization as a recommendation task which
attempts to assign, to each user impression, the variant that a user is most
likely to click on. 
Note that once the ad variant has been selected, the resulting ad impression
looks much like non-dynamic ad. Figure~\ref{exampleYS} shows an instance of a
dynamic ad campaign for Yahoo! Shopping.

Display ads, as opposed to content (e.g. articles, search results) and other
digital advertising types (e.g. sponsored search) are characterized by very low
CTR. In addition, many campaigns {\em target} (i.e. compete over) the same set
of users, resulting in most users being exposed to a campaign no more than once. 
These two factors lead to an extremely sparse user signal - most users do
not register a single click in most campaigns - and impose a great challenge
to any recommendation algorithm.

As there may be returning users observed across campaigns, one might consider
the possibility of cross-campaign modeling. However, the differences in granularity of considerations
within and between campaigns make cross campaign
modeling extremely challenging. This stems from the need to distinguish, within each campaign,
between preferences among relatively similar items, while different campaigns
may apply to very different and uncorrelated commercial domains. For example, it
is not likely that similar movie preferences of two users will imply much
about the resemblance of their preferences in rental cars. Thus, this work
focuses on standalone dynamic campaign optimization. 

\section{Notations}
\label{problemDescription}
Our problem is about recommending items, in our case ad variants ${\bf A} =
\{a_1,a_2,\ldots,a_L\}$, to users ${\bf U} = \{u_1,u_2,\ldots,u_M\}$.
Each user $u_i$ is associated with a set of feature values     
$[u_{i}^1,u_{i}^2,$ $\ldots,u_{i}^K]$, where $u_{i}^k \in {\bf F}_k$, and ${\bf
 F}_k$ is the set of available values for feature $k$
(e.g., one feature can be `gender', where the set of available values is
`male', `female' and `unknown').

The set of observations constructing the training data is denoted by $\bf
T=\{t_d\}_{d=1}^P$ (where $P \approx M$). Each observation is a 3-tuple $t_d=\paren{u_i, a_j, c_d}$,
that includes a user, an ad variant and a binary reward - either a positive reward (i.e. click) or 
a negative reward (i.e. non-click).

We now define $\bf C$ as the set of (user, ad variant) pairs that resulted in a
click ${\bf C}=\left\{(u_i,a_j) \left.\right| (u_i,a_j,c_d) \in {\bf T}, c_d =
click\right\}$. Similarly, we denote by ${\bf NC}$ the set of (user, ad variant)
pairs that did not result in a click. Finally, let ${\bf N}={\bf C}\bigcup{\bf
NC}$ be the union of the two sets.

\section{The OFF-Set Algorithm}
\label{solution}
Dynamic ad selection is yet another ranking problem. Given a new user entering the system, we aim at selecting the best possible ad variant that will maximize the probability of a click, thus maximizing the CTR of the overall campaign. Like in many other ranking problems, we chose to use a latent factors 
approach, where each ad variant and each user are represented by a latent vector, ${\bf v}_{a_j}\in R^D$ and ${\bf v}_{u_i}\in R^D$ respectively. The score of the match for a pair $(u_i,a_j)$ is defined by the inner product of the two corresponding vectors,
$S(u_i,a_j) = \left\langle {\bf v}_{u_i}, {\bf v}_{a_j} \right\rangle =
{\bf v}_{u_i}^T\cdot {\bf v}_{a_j}$, where higher inner products represent
better match, or a higher probability of a click. Hence, given a user, the
OFF-Set algorithm suggests the ad variant that results in the highest matching score, in
order to maximize the CTR of the campaign. As previously mentioned, the user
vector is not directly trained by the algorithm, but rather the feature vectors
that construct it, as will be detailed in section \ref{userCombination}.

\subsection{Applying Maximum Log Likelihood\label{constraint}}
We define the target function for maximization following the approach suggested by Aizenberg et al.
\cite{radioActive}. Given the training data described above, that includes
$\abs{\bf C}$ click interactions, we define a multinomial distribution for a
click over all our (user, ad variant) pairs as follows:   
\begin{eqnarray}
PC(u_i,a_j) = {\abs{\bf C}}\cdot\frac{\exp^{S(u_i,a_j)}}{\sum_{(u_k,a_l)\in {\bf N}}\exp^{S(u_k,a_l)}},
\end{eqnarray}

where $PC(u_i,a_j)$ is the probability of a click for the $(u_i,a_j)$ pair, and
$S(u_i,a_j)$ is the score our model assigns this pair. Note that this is a
general definition of probability that satisfies $\sum_{(u_i,a_j)\in {\bf N}}
PC(u_i,a_j) = \abs{\bf C}$.
Also, it holds that for all $i$ and $j$, $0 < PC(u_i,a_j) \leq 1$ under the constraint that
$\abs{S(u_i,a_j)} \leq \ell$, for\footnote{As the score value is not important by itself, but
rather the ranking it derives, scores can always be normalized in order to
satisfy this constraint. Therefore, from now on, we will not refer to this
constraint.} 
\begin{eqnarray}
\ell=\paren{0.5\ln{\frac{\abs{\bf N}}{\abs{\bf C}}}}.
\end{eqnarray} 
We prove this by,
\begin{eqnarray}
\frac{\abs{\bf C}\cdot \exp^{S(u,p)}}{\sum_{(u_k,a_l)\in {\bf N}}\exp^{S(u_k,a_l)}} \leq \frac{\abs{\bf C}\cdot \exp^\ell}{\abs{\bf N}\exp^{-\ell}} =  \frac{\abs{\bf C}}{\abs{\bf N}}\exp^{2\ell} = 1.
\end{eqnarray}
Our model's parameters (denoted hereafter by $\Theta$) are now trained in order
to maximize the mutual probability of the pairs that actually resulted in a
click (assuming independence between all pairs). Using the log likelihood
approach, we get
\begin{eqnarray}\label{cost1}
\Theta &=& \arg\max \log \prod_{(u_i,a_j)\in C} PC(u_i,a_j) \\ \nonumber
&=& \arg\max \sum_{(u_i,a_j)\in C} \log PC(u_i,a_j)\\ \nonumber
&=& \arg\max \sum_{(u_i,a_j)\in C}  \left( \log \abs{\bf C} + \log \exp^{S(u_i,a_j)}\right.\\ \nonumber
&&\left.- \log \sum_{(u_k,a_l)\in {\bf N}}\exp^{S(u_k,a_l)}\right) \\ \nonumber
&=&\arg\max \left(\sum_{(u_i,a_j)\in C} S(u_i,a_j) + \abs{\bf C}\log \abs{\bf C}\right.\\ \nonumber
&&\left.- \abs{\bf C}\cdot\log\sum_{(u_i,a_j)\in {\bf N}} \exp^{S(u_i,a_j) }\right) .
\end{eqnarray}

\subsection{Training The Model}
We extract $\Theta$ using a stochastic gradient ascent method. The derivative
of the above target function depends on whether the (user, ad variant) pair
resulted in a click or not. Whenever $(u_i,a_j)\in{\bf C}$, we get:
\begin{eqnarray}
\Delta\Theta &=& \frac{\partial S(u_i,a_j)}{\partial \Theta} - PC(u_i,a_j) \frac{\partial S(u_i,a_j)}{\partial \Theta}\\
&=& (1-PC(u_i,a_j))\cdot \frac{\partial S(u_i,a_j)}{\partial \Theta}.
\end{eqnarray}
When $(u_i,a_j)\in{\bf NC}$, the first component is eliminated, and we get
\begin{eqnarray}
\Delta\Theta &=&  - PC(u_i,a_j) \cdot\frac{\partial S(u_i,a_j)}{\partial \Theta}.
\end{eqnarray}

It is clear that a positive reward (a click) will always result in an increase in the value of $S(u_i,a_j)$, and vice versa. However, the increase and decrease step sizes are not equal.

If we begin the training process with random variables, assuming 
equal probability of a click to all pairs (which equals $\frac{\abs{\bf
C}}{\abs{\bf N}}$), we get that the ratio (denoted by $\mu$) between the step sizes of a negative reward and a positive reward is:
\begin{eqnarray}\label{ratio}
\mu = \frac{-\frac{\abs{\bf C}}{\abs{\bf N}}}{1-\frac{\abs{\bf C}}{\abs{\bf N}}} = -\frac{{\abs{\bf C}}}{{\abs{\bf N}-\abs{\bf C}}} = -\frac{\abs{\bf C}}{\abs{\bf NC}}.
\end{eqnarray}
This implies that for any positive reward we should apply a gradient ascent
step that will increase $S(u_i,a_j)$ in some learning step $\alpha$, while for any negative reward, we apply a gradient ascent step of size $\alpha\cdot\mu$, that will decrease $S(u_i,a_j)$.

Throughout the training process, assuming the model succeeds in representing the
user preferences by representing the real probabilities, the step sizes should
be decreased in both directions, and dependent on the specific pair. For simplicity reasons, we keep $\mu$ as the
ratio between the update steps, and update its value throughout the training
process, as detailed in Section~\ref{subsec:online}. 

\subsection{Alternative Definition of Target Function}
Having first derived the learning procedure from a
probabilistic viewpoint, we now present an equivalent, simpler and more
intuitive target function that derives the exact same learning procedure. We find
the equivalence between the two settings (derived from quite different
approaches) interesting by itself.

As ranking the best ad variants per user is our main goal, we might ask to
simply maximize the score our model assigns to a positive reward in the data,
compared to the score the model assigns to a negative reward. We formulate this
by the following target function:    

\begin{eqnarray}\label{cost2}
\Theta = \arg\max &&\frac{1}{\abs{\bf C}}\sum_{(u_i,a_j)\in {\bf C}} S(u_i,a_j) -\\ \nonumber
&& \frac{1}{\abs{\bf NC}}\sum_{(u_i,a_j)\in {\bf NC}} S(u_i,a_j)\\ \nonumber
&& s.t. \forall{{\bf v}_i\in\{{\bf v}_{u_i},{\bf v}_{a_j}\}}_{u_i\in {\bf U}, a_j\in {\bf A}}\left\|{\bf v}_i\right\|_\infty \leq b,
\end{eqnarray}
where $b$ is a scaling factor that limits the $L_\infty$ norm of all vectors. 
This factor is the counterpart of the constraint on the score function dictated in Section \ref{constraint}.

Again, we apply the stochastic gradient ascent method. For a $(u_i,a_j)$ pair that resulted in a click we get

\begin{eqnarray}
\Delta\Theta = \frac{1}{\abs{\bf C}}\frac{\partial S(u_i,a_j)}{\partial \Theta},
\end{eqnarray}
and for a negative reward we get
\begin{eqnarray}
\Delta\Theta = -\frac{1}{\abs{\bf NC}}\frac{\partial S(u_i,a_j)}{\partial
\Theta}.
\end{eqnarray}
We then scale our vectors to obey the norm constraint (when multiplying all
vectors in the same factor, the ranking order does not change).
As expected, the direction of updates are opposite (a positive reward results in
an increase of the score and vice versa). Furthermore, the ratio between a
negative and a positive reward updates is exactly $\mu$.

\subsection{Online Training}
\label{subsec:online}
Up until now we have referred to the training data as a fixed and known set of
examples. 
This, however, does hold in the real world settings we operate in. The amount of
data pairs is large, and the rate in which they are produced is high. On the one
hand, the data sparsity issue requires gathering many samples for training.
On the other hand, as a single campaign may run for only a few hours,
the training must be done with minimal latency. Considering
these constraints together, an on-line training scheme is required. In such
a solution, each sample updates the model's parameters only once (either
positively, or negatively, according to the reward). Furthermore, the update
depends only on this single sample, with no reliance on previous data samples.
The only accumulative value we aggregate from the collection of data samples is
the updated $\mu$ factor, the ratio between the step sizes of updates stemming
from positive and negative rewards.

Specifically, OFF-Set computes an updated value of $\tilde{\mu} = \frac{\abs{\bf
C}}{\abs{\bf NC}}$ every $1000$ ad impressions, and smoothes it into the $\mu$
value by $\mu=\gamma \cdot \tilde{\mu} + (1-\gamma)\mu$, where $\gamma$ in our
experiments is $0.02$. Such a continuous update of $\mu$ enables adaptation to a
new value, when the reward trend changes.     

\subsection{OFF-Set Adaptation to Trends}
\label{subsec:trend-adapt}
Users' preferences often change over time. This may happen due to several
reasons; the nature of surfing is different in different times of the day or the
week (at home vs. at work, day time vs. night time, working days vs. weekends,
etc.). Moreover, some variants may become more or less attractive over time
(relevancy to user changes). Also, some campaigns may add or omit products
throughout the lifespan of the campaign, changing the set of items to be
ranked. Therefore, any algorithm that handles users' online preferences must
address trend changes.        

Although the target functions presented in Equations \ref{cost1} and \ref{cost2}
are both trend agnostic, the fact that OFF-Set scans the data in a single,
temporally-ordered pass rather than in random order across multiple iterations
causes its learned model to be more influenced by the latest observations than
by earlier ones.

This works in our favor in the online setting\footnote{OFF-Set is thus less
suited for random train-test splits, where training examples are not
presented by temporal order and when test points do not temporally follow the
training examples.}. 
When there is no change in trend, the model converges into a local maxima,
while observations that appear after a trend change will result in a convergence
to a different solution. This feature is demonstrated in our evaluation section. 

\pagebreak
\section{User and Item Latent Vectors}
\label{userCombination}
We now turn to describe the separate latent vector representations that map
the ad variants and users into the same latent space. 

For ad variants, we typically have no
features, or rather any features we may have are common
to all variants as they are associated with the entire campaign, and cannot serve
to distinguish between the variants. Yet, the variants
are more stable and do not suffer from the perpetual cold start
problem exhibited by the users. Therefore, we simply assign one $D$-dimensional
latent factor per ad variant (item) $a_j$, ${\bf v}_{a_j} \in R^D$.

We offer a novel construction of the latent representation on the user side.
Recall that we assume that users are encountered once, and so are perpetually
cold. Hence, they must be modeled by their features. Previous works in matrix
factorization assigned a latent vector to each feature value, and modeled
users (or items) as a linear combination of their features' vectors
\cite{radioActive,C2C-WWW2012}.
While this approach can handle sparse user data, it cannot capture any
dependencies between the features as the combination between the features is
linear.
\label{subsec:dependentlatentspace}
Instead, we construct a more complex latent space over user features, that
allows the representation of strong dependencies between each pair of features.

Recall that items were mapped into $R^D$, and assume that each user has some
value across $K$ features. We select an {\em overlap dimension} $o$ and a {\em
standalone dimension} $s$ such that $D=Ks + {K \choose 2}o$. Each feature value
is assigned a $d$-dimensional vector where $d=s+(K-1)o$, so that ${\bf v}_{u_i}^k \in R^d$ is the vector assigned to the $k$th feature value of user $u_i$. That vector has $s$
entries that are attributed only to the specific feature, and $K-1$ blocks
of $o$ entries that are in overlap with each of the $K-1$ other
features\footnote{For simplicity, we assume each feature merits $s$ unique
entries and  every pair of features merits an overlap of $o$ entries.
Non-uniform assignments of entries are possible and have no effect on the rest
of the algorithm.}.
 
\begin{figure}
\centering
\epsfig{file=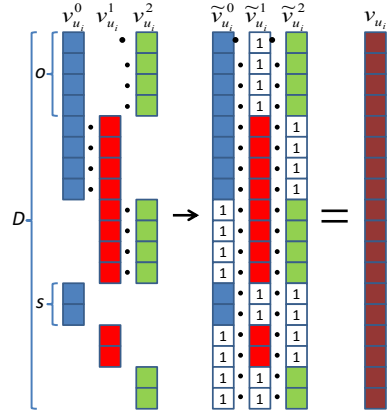, width=0.99\columnwidth}
\caption{Example of a user latent factor construction from its feature value vectors ($K=3$). In
this example, $D=18$, $d=10$, $s=2$ and $o=4$. The middle illustration presents the auxiliary vectors $\widetilde{\bf v}_{u_i}^k$\label{userVectorConstruction}}
\end{figure}

We now compose ${\bf v}_{u_i}\in R^D$ from ${K}\choose{2}$ chunks of size $o$,
each holding the products of the values from a pair of feature vectors. The
rest of ${\bf v}_{u_i}$ holds the $Ks$ standalone values from each of the
feature vectors. The process is illustrated in Figure \ref{userVectorConstruction}. 
The specific indices are fixed per each feature and feature pair, and as a
whole, obey the following criteria:     
\begin{itemize}
\item Every two features share $o$ distinct entries from $1,\ldots,D$.
\item $s$ entries from $1,\ldots,D$ are dedicated to each feature alone.
\end{itemize}
Now, let $\widetilde{{\bf v}}_{u_i}^j \in R^D$ be an extension
of ${\bf v}_{u_i}^j$, so that the values of ${\bf v}_{u_i}^j$ appear in $d$ of
the entries of $\widetilde{{\bf v}}_{u_i}^j$, as per the above criteria, and the 
rest of the entries are set to $1$ (see middle part of Figure \ref{userVectorConstruction}). 
Using this formulation, ${\bf v}_{u_i} = \prod_{1\leq j\leq K} \widetilde{{\bf v}}_{u_i}^j$ (for an element-wise product).

Note that in settings where items are also associated with features, one can
trivially extend the methodology of our user-side construction to the item
side.

After describing the structure of the latent spaces we use, the training
of the different latent spaces is fairly straightforward. Each training step
requires an increase or decrease in the score function, defined by an inner
product of the user vector and the ad variant vector. Such a change includes two
stages. First, the ad variant vector is shifted in the direction of the gradient
(which is the combined user vector). Then, 
we update each of the feature value vectors associated with the user.
For each such update we compute the gradient vector, combined from the ad
variant vector, and the other feature value vectors of the user.         

\section{Evaluation}
\label{results}
We evaluate OFF-Set against two state-of-the-art methods mentioned in Section
\ref{sec:related} as well as a simple `popularity' algorithm. We measure
algorithms recommendation quality on two sets of synthetic (model generated)
data, and on two sets of real-life offline data.

This section is structured as follows. We
describe the adaptation of the baselines to our experiments in Section
\ref{baselines}. We discuss the evaluation metric in Section
\ref{evaluationmetric}. Sections \ref{synthetic} and \ref{realdata} then
describe the datasets and present results for both the synthetic and real-life
datasets, respectively.

\subsection{Baselines\label{baselines}}
We use four baselines throughout our experiments: two flavors of Factorization
Machines (FM) \cite{FM,FM2}, Gradient-Boosted Decision Trees (GBDT)
\cite{GBDT1,GBDT2} and a global-popularity based algorithm (Popularity).

\begin{figure}
\centering
\epsfig{file=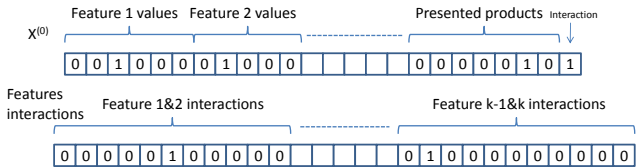, width=0.99\columnwidth}
\caption{Single data entry for running the Factorization Machine algorithm,
without paired user features interactions (top), and with paired user features
(bottom).\label{representationLikeFM}}
\end{figure}

\textbf{Factorization Machines} present a framework for
modeling sparse feature spaces, where each data sample is encoded as a real
valued feature vector with some target. In our experiments, all features are
categorical and the target is a binary reward. We thus build a binary
vector with a binary target as input for FM.
Each possible user feature value and each ad variant are associated with a
specific index in the input vector, as illustrated in upper part of Figure
\ref{representationLikeFM}. The indices corresponding to the feature values and
ad variant in each specific observation $t_d$ are lit, and the reward bit is
set accordingly.

In a straightforward adaptation of the data to FM, a latent factor is learned
for each of the user feature values and each ad variant. Thus, a second order FM,
reflects dependencies between every two vector entries (one of which
may represent an ad variant and the other a user feature value). However,
dependencies between two user features and an ad variant cannot be reflected in
this manner. For this purpose, and in order to allow for a fair comparison
against OFF-Set, we introduce {\bf FM with paired user features}. This variant
of FM, also used as a baseline, uses longer feature vectors with additional
bits representing the possible interactions of all pairs of user
feature values (see lower part of Figure \ref{representationLikeFM}).
This allows FM to model pairwise user feature dependencies and produces an
equivalent solution to a $3$rd order FM for the problem at hand (a similar
approach was presented in \cite{FM2}).

Our experiments use the {\em libFM} source code\footnote{http://libfm.org/},
with the Markov Chains Monte Carlo configuration solving a classification problem.

\textbf{GBDT} was applied by using the {\em Simple GBDT} source
code\footnote{http://code.google.com/p/simple-gbdt/}, providing it with the
input data in the binary vector representation as described in the top
part of Figure \ref{representationLikeFM}. 
Here, there is no need for adding indicators for user feature
interaction, as these can be captured by the model.

\textbf{Popularity} is a simple yet powerful baseline that does not perform
personalization at all. Rather, it models the ad variants' general popularity.
For each user, it suggests the list of ad variants in a descending order of
historical CTR. This algorithm follows the trend of the data, by applying a decay factor
to the CTR values of all variants every 1M observations, in order to increase
the effect of recent activity.    

\subsection{Evaluation metric\label{evaluationmetric}}
Our goal is to measure the effectiveness of various recommendation schemes in
serving ad variants to users, based on offline data as described in
Section~\ref{problemDescription}, where each observation $t_d=\paren{u_i, a_j,
c_d}$ encodes the action (click or lack thereof) of a certain user upon seeing a
certain ad variant. 
Using such data, we simulate a serving scenario where each algorithm, given
an observation $t_d$, ranks the ad variants by descending match order for
the user encoded in the observation.

Consider an observation $t_d \in {\bf C}$; a tested algorithm may not have
recommended (i.e. ranked in the top-most position) the same ad variant that was actually
served to the user and resulted in a click. However, unlike rejection sampling
techniques such as proposed in \cite{lihongli:contextualbandits}, and since
clicks are a rare commodity in display advertising, we still want to assign some
score to the algorithm's  ranking based on this observation. Intuitively, the
higher the algorithm ranked the clicked ad variant, the better. Concretely, we
utilize Mean Reciprocal Rank (MRR), a popular metric in many ranking problems
\cite{baeza2011modern}, and score the algorithm according to the reciprocal rank
it assigned to the clicked ad variant. 

To complete the picture, consider an observation $t_d \in {\bf NC}$. Since
non-clicks are by far the more frequent outcome in display advertising, even
when the ad is well targeted to the user, we do not penalize our tested
algorithms even if they wished to serve the same ad variant that was not
clicked. Thus, we do not score algorithms with observations from ${\bf NC}$, and 
measure MRR using only the set ${\bf C}$ of clicked observations. Formally,  for
algorithm $A$,
\begin{eqnarray}
mrr(A)=\frac{1}{\abs{\bf C}} \sum_{(u_i,a_j)\in{\bf C}}
\frac{1}{r_A^{u_i}(a_j)},
\end{eqnarray}
where the function $r_A^{u_i}(a_j)$ is the rank of the ad variant $a_j$ in
algorithm $A$'s computed ranked list for user $u_i$ ($1$ for the first
location). Under the assumption that the presented ad variants appear
uniformly at random in the test data (and independently of the user), this
metric quantifies well the ability of an algorithm to predict a click.

\subsection{Evaluation on Synthetic Data \label{synthetic}}
Synthetic data used in our evaluation is generated by a pre-defined set of
rules. The generator allows setting the number of ad variants, any
combination of user features that imply a preference to any variant,
and the CTR lift of such preferences over a default value.  

We perform two synthetic experiments: 
\begin{itemize}
\item Stable data - preferences do not change over time.
\item Trending data - preferences change over time.
\end{itemize}

\subsubsection{Stable Data}
This experiment sets the number of ad variants to five, and uses the rules
described in Table \ref{synRules}.
The CTR lifts following these rules are accumulative. That is, a
user-ad variant combination will accumulate CTR for every rule they
satisfy. For example, CTR for people in California on ad variants $0$ or $1$ is
$0.001$, while on variant $2$ it is $0.011$. For a person in New York, born
in 1953, the click probability on ad variant $1$ is $0.301$.
As the user profiles are randomly generated, and the ratio of people whose
features fit rules $3-6$ is small, it is clear that the best ad variant is $2$.
Yet, an algorithm that can identify certain users who prefer ad variants $0$ or
$1$ is likely to win.   
\begin{table}
\center
\begin{tabular}{ccccc}
\hline
Age&Geo&Gender&Ad&Accumulative\\
& & & & CTR lift\\
\hline
ALL & ALL & ALL & ALL & 0.001\\
ALL & ALL & ALL &  2 & 0.01\\
$1980-1989$&New York&ALL&0&0.30 \\
$1950-1959$&New York&ALL&1&0.30 \\
$1980-1989$&Arizona&ALL&1&0.30 \\
$1950-1959$&Arizona&ALL&0&0.30 \\
\hline\\
\end{tabular}
\caption{Rules for generating synthetic click data.\label{synRules}}
\end{table}

Using the above configurations, we generate training as well as test data. Each
sample requires generation of a user, with random features, and a random ad
variant. Then, according to the generated pair, the expected CTR according to
the model is computed, and based on this probability, the binary reward is set.

Training data consisting of $8$ million impressions was generated for this
experiment. For FM (both flavors) and GBDT, the training data was further
sampled to include all clicks and two random non-clicks per click\footnote{Tests
with non-sampled training data for FM resulted in poor results (MRR of
$0.719$), while for GDBT training time on all 8M samples was too high.}.

Then, MRR was computed for OFF-Set and each baseline algorithm on test data
consisting of $8$ million additional impressions, that included $26,905$ clicks (only $3,180$ of which
were due to rules $3-6$). Results are reported in Table~\ref{synRes}.
For the comparison, we used the best results we were able to achieve for each of
the FM/GBDT versions using different configurations and varying number of
training iterations.
According to Hoeffding's inequality \cite{Hoeffding}, computed with confidence of $95\%$, a
gap of $0.016$ between two MRR values reflects a statistically significant
difference between the underlying algorithms.

As reported in Table \ref{synRes}, OFF-Set, GBDT and FM with paired
user features, achieve similar results. They outperform ``vanilla'' FM and
Popularity. We note that OFF-Set achieved these results in a purely on-line
manner, and required no iterations over the training data. FM
, on the other hand, is able to achieve result comparable to OFF-Set only after
a minimum of $50$ iterations.

\subsubsection{Trending Data}
As argued in Subsection~\ref{subsec:trend-adapt}, trends in user preferences may
change over time. Trend changes are not known in advance and can be
difficult to detect in real time.
The second synthetic experiment illustrates the effect of such trend
changes in the data. 
To simulate trend changes, we created a synthetic dataset in which the first 4M
training examples were again generated according to the model of Table
\ref{synRules}. Then, 4M additional training examples were generated
similarly, with one difference: the most popular item changed from being
number $2$ to being number $3$ (the second rule was updated). After training on
these 8M examples (with subsampling of non-clicks for GBDT and both FM variants,
as explained above), the algorithms were tested on 8M test observations following the new trend. The results
are presented in the right-most column of Table \ref{synRes}.
Here we can see a clear advantage of OFF-Set, compared to both FM and GBDT. The
Popularity algorithm, being adaptive to trend changes, preserves its former performances.

\begin{table}[tb]
\begin{tabular}{|l|c|c|}
\hline
Algorithm & Stable data & Trending data \\
& MRR results& MRR results\\
\hline
OFF-Set &  0.8389 & 0.8231\\
Popularity &0.7719 & 0.7667\\
FM & 0.8051  & 0.5980\\
FM w/paired features &  0.8390 & 0.6803 \\
GBDT & 0.8327 & 0.6387  \\ 
\hline
\end{tabular}
\caption{MRR results on both stable (middle column) and trending (right
column) synthetic data.\label{synRes}}
\end{table}

\subsection{Evaluation on Real Ad Campaign Logs\label{realdata}}

\begin{table*}[tb]
\centering
\begin{tabular}{|l|c|c|c|c|}
\hline
Campaign & clicks in & clicks in & number of & Significant gap\\
 &warm-up phase & online test phase & ad variants & in mrr \\
\hline
Campaign A &  100 & 2400 & 14 & 0.055\\
Campaign B &  100 & 900 & 22 & 0.091\\
\hline
\end{tabular}
\caption {Real-life campaign statistics. Significance was computed from Hoeffding's formula with confidence of $95\%$.\label{realData}}
\end{table*}

For evaluation on real data we chose two different dynamic campaigns, henceforth
referred to as Campaign A and Campaign B.
The target audience for both campaigns, as defined by the advertisers,
consisted of two geographic locations, two genders and five age ranges - 20 possible user
profiles overall.

In order to simulate an on-line setting, the test (on both campaigns) was
performed as follows:
\begin{enumerate}
\item First, some initial number of observations is used for warm-up training.
\item Subsequently, all remaining observations are iterated over,
sorted by time of appearance. Observations that did not result in a click are
used for further training. Any observation that resulted in a click is first
used for evaluation, by requesting each algorithm to rank the available items
for the given user. Afterward, the observation (along with its positive
reward) is added as an additional training example.
\end{enumerate}
Table \ref{realData} provides some statistics on both campaigns, including the
number of clicks contained in the warm-up and online test phases described
above. Due to business sensitivity, we cannot disclose the exact
number of non-clicks in the data. Rather, we roughly disclose that in both
campaigns, the number of non-clicks was 2-3 orders of magnitude larger than the
number of clicks.

For OFF-Set, being a single pass algorithm, the procedure above is
straightforward. FM and GBDT, however, must be retrained from scratch once
encountering a click in the online portion of the test\footnote{Since MRR is
only measured on clicks, there is no need to retrain these models, for the
purposes of the experiment, on each non-click.}. 

Having established the superiority of FM with paired user features over the
vanilla version, we omit vanilla FM from the evaluation here. Instead, we add a
{\em random} baseline, which ranks ad variants by a random permutation in any
observation.
The results are presented in Table \ref{realRes}.
We can see a clear and significant advantage to OFF-Set as compared to the
baselines (see Table \ref{realData} for the significant gap values for each campaign). In addition, its efficiency, being a single pass method, along with
its low memory footprint, establish its superiority over the other methods for
this problem.

\begin{table}[tb]
\centering
\begin{tabular}{|l|l|l|}
\hline
Algorithm & MRR results & MRR results \\
 & campaign A & campaign B \\
\hline
Random     & 0.2114 &0.2255\\
Popularity & 0.2940 &0.4360\\
FM w/paired features &  0.3038 & 0.3824\\
GBDT & 0.2613  &0.3193\\
OFF-Set & 0.5993&0.6234\\
\hline
\end{tabular}
\caption{MRR results on real (offline) data. MRR differences higher than $0.055$ and $0.091$ for campaigns $A$ and $B$ respectively are significant (with 95\% confidence).
\label{realRes}}
\end{table}

\section{Conclusions and Future Work}
\label{sec:conc}

In this work we introduced OFF-Set - an online single-pass algorithm that handles a perpetual user cold start problem. OFF-Set is based on Matrix Factorization, and is motivated by a log-likelihood approach. It exploits repeating user features 
and captures pairwise feature dependencies.
We evaluated OFF-Set in the context of dynamic ad optimization on both synthetic, model-generated,
and real-life data, and demonstrated its superiority over state-of-the-art methods. 

\pagebreak
The following directions are left for future work:
\begin{itemize}
\item Investigate the value of modeling users across campaigns in similar
commercial domains (advertizing verticals).
\item Whenever features of ad variants are available, introduce them into
the model in order to combat click sparsity on items. This may prove
particularly beneficial in campaigns exhibiting a large number of
ad variants.
\item Perform quasi-hierarchical smoothing of latent vectors of features, where
applicable. Examples include smoothing the vectors of neighboring age ranges and
neighboring geographical regions.
\item Investigate the applicability of Contextual Bandits algorithms \cite{
langford:contextualbandits} to the dynamic campaign optimization problem, with
context being the user attributes.
\end{itemize}

\bibliographystyle{abbrv}
\bibliography{bibfile}  
\end{document}